\documentclass{article}

\usepackage[T1]{fontenc}
\usepackage[utf8]{inputenc}
\usepackage{color}
\usepackage[brazilian,english]{babel}
\usepackage{times}
\usepackage{anyfontsize}

\usepackage{amsmath}
\usepackage{amssymb}
\usepackage{amsfonts}

\usepackage{graphicx}
\usepackage{psfrag}
\usepackage{subcaption}

\usepackage{amsthm}
\usepackage{thmtools}
\usepackage{thm-restate}

\usepackage{tikz}
\usepackage{pgfplots}
\usepgfplotslibrary{patchplots}
\usetikzlibrary{fit,positioning}

\usepackage{xspace}
\usepackage{multirow}
\usepackage{mathtools}
\usepackage{cancel}
\usepackage{footnote}

\usepackage{iclr2016_conference}
\usepackage{hyperref}
\usepackage{url}

\title{Reducing the Training Time of Neural\\Networks by Partitioning}

\author{Conrado S. Miranda \& Fernando J. Von Zuben \\
School of Electrical and Computer Engineering \\
University of Campinas, Brazil \\
\texttt{contact@conradomiranda.com, vonzuben@dca.fee.unicamp.br}
}

\tikzstyle{box}=[rectangle, draw=black!100]
\tikzstyle{neuron} = [circle, inner sep=0pt, minimum size = 1.5mm, draw = black!100, fill = white!100]
\tikzstyle{textneuron} = [circle, inner sep=2pt, draw = black!100, fill = white!100]

\begin{document}

\maketitle

\begin{abstract}
  This paper presents a new method for pre-training neural networks that
  can decrease the total training time for a neural network while maintaining
  the final performance, which motivates its use on deep neural networks.
  By partitioning the training task in multiple training subtasks with
  sub-models, which can be performed independently and in parallel, it is shown
  that the size of the sub-models reduces almost quadratically with the number
  of subtasks created, quickly scaling down the sub-models used for the
  pre-training.
  The sub-models are then merged to provide a pre-trained initial set of weights
  for the original model.
  The proposed method is independent of the other aspects of the training, such
  as architecture of the neural network, training method, and objective, making
  it compatible with a wide range of existing approaches.
  The speedup without loss of performance is validated experimentally on MNIST
  and on CIFAR10 data sets, also showing that even performing the subtasks
  sequentially can decrease the training time.
  Moreover, we show that larger models may present higher speedups and
  conjecture about the benefits of the method in distributed learning systems.
\end{abstract}

\section{Introduction}
Research in deep learning using neural networks has increased significantly over
the last years. This occurred due to the ability of deep neural networks to
achieve higher performance when compared to other methods on problems with a
large amount of data \citep{Bengio-et-al-2015-Book} and advances in computing
power, such as the use of graphic processing units (GPUs)
\citep{raina2009large}.

Despite the advances obtained by using GPUs for training deep neural networks,
this step still can take a lot of time, which affects negatively both research
and industry as new methods take longer to be tested and deployed. Some
researches have focused on speeding up deep neural networks in general,
including proposals based on hardware, such as using limited numerical precision
\citep{gupta2015deep}, which could increase the number of computing units on the
hardware, and software, such as using Fourier transform to compute a convolution
\citep{vasilache2014fast}. These and other methods optimized for computation of
neural networks on GPUs lead to the development of domain-specific libraries,
such as cuDNN \citep{chetlur2014cudnn}. In this paper, we focus on existing
research interested in decreasing the training time, as these approaches are
closer to the proposed method. However, we highlight that these improvements are
not mutually exclusive and can be used together.

\citet{Krizhevsky14} proposed a mixture of data and model parallelism over GPUs
in a single machine based on the type of the layer, exploiting their
particularities for increased speed. Essentially, the convolutional layers
exploit data parallelism, since they are the most computing intensive part of
the neural network, and the fully-connected layers exploit model parallelism,
since they have most of the parameters and may not fit in a single GPU. This
leads to a significant speedup in comparison to other existing methods for
training convolutional neural networks over a GPU cluster.

DistBelief \citep{dean2012large} is another framework to speedup the training of
large neural network by exploiting parallelism, but it focuses on clusters of
computers. The data parallelism is exploited by dividing the data set in shards
that are fed to different groups of computers, where each group replicates the
full model and are synchronized through a parameter server. For the model
parallelism, a locally connected neural network is used in order to reduce the
communication among machines that jointly represent one replica of the model.
Since the model is locally connected, only the edges that cross the machine
boundary need to be transmitted. This framework is extended in
\citet{coates2013deep} to use GPUs as computing units.

These methods of model parallelization to handle large neural networks require
communication between the multiple computing units, which usually is slower than
the computation and characterizes an overhead in the learning process. Other
methods of parallelization, such as computing the branches of a model like
GoogleNet \citep{googlenet} in parallel, also require communication among the
units. These communications are proportional to the size of the network, hence
presenting less overhead for smaller models.

Moreover, even if a single GPU can hold the full model and the model is dense
and occupies all the computational resources, all computations of a layer may
not be performed in parallel, leading to an increase in test and training time.
For instance, VGGNet \citep{vggnet} is a very dense model and may be able to
keep the GPU completely busy. But consider the two fully-connected layers with
4096 in this network. If the GPU has 4096 cores and we consider only the time
for multiplications, a total of 4096 GPU cycles will be required per sample,
while only 1024 cycles are required for a model with half as many neurons. So
smaller versions of dense models are beneficial even if we discard the
associated overheads.

Therefore, using smaller models can speedup the task, but smaller models may not
be able to achieve the same performance attainable by larger models. However,
the learning process does not have to adjust the full model in all iterations.
Pre-training methods exploit this characteristic by providing adjusted, instead
of fully random, initial conditions for the training to continue with the full
model.

A well known example of this is unsupervised pre-training \citep{erhan2010does},
where the network is greedly trained layer-wise to reconstruct its input. Since
only the parameters of the current layer have to be learned, this pre-training
requires less memory and processing time to fit the model. After the initial
iterations of pre-training, the full model is built stacking the layers on top
of each other and the training continues, performing a fine-tuning of the
parameters found. However, two important limitations of unsupervised
pre-training are that it requires a generative model of the data to be known,
which can be different from the desired task, and that the layers must be learnt
sequentially.

In this paper, we introduce a new method of partitioning the network to perform
a pre-training. Instead of partitioning layer-wise, the proposed method
partitions the neural network in smaller neural networks whose tasks are
equivalent to the original training task, avoiding the need to create substitute
tasks such as generative costs, for pre-training. Moreover, the subtasks are
independent of each other, which allows them to be learnt in parallel without
communication. Another advantage of the proposed method is that, since the task
is kept the same, the learning algorithm used to adjust the parameters can also
be the same for all stages. Therefore, it can be viewed as a higher level method
and is compatible with existing training strategies. After the proposed
pre-training is complete, the obtained smaller neural networks are merged and
used as initial condition for the original neural network.

The new method is also straightforward to implement and decreases the number of
parameters of the subtasks quadratically in the number of subtasks created, thus
being characterized as a highly scalable approach. We perform two experiments,
one with MNIST and another with CIFAR10 data sets, to show that the training
time required is indeed reduced while the generalization capability may not be
affected by the new method. Furthermore, the experiments also show that the
method can be used to speedup the training even when the subtasks are performed
sequentially, that larger models have higher speedups, and that the sub-models
learn different representations for the data, which is advantageous for the
performance when merging these sub-models.

Recently, an approach called Net2Net \citep{net2net}, which is similar to the
one proposed in this paper, was made published. Both approaches were developed
in parallel by different research groups and focus on speeding-up the training
of neural networks while maintaining the same performance level achievable when
training the full model from a random initialization. However, Net2Net requires
the existence of a pre-trained neural network and focuses on training a larger
model faster than from a random initialization by using the existing network as
initial condition, while our method does not require any pre-trained network, as
the pre-training is part of the method. Therefore, the speed-up analysis we
perform for our method includes both the pre-training and the regular training,
while the analysis presented by \citet{net2net} does not include the training
time of the teacher network. In fact, it is possible that the total training
time of the teacher network plus the expanded network is higher than the
expanded network directly. We compare our method to Net2Net in the CIFAR10
experiment.

This paper is organized as follows. Section~\ref{sec:partition} provides the
motivation, description and main advantages of the method proposed.
Section~\ref{sec:experiment} describes the experiments performed to test the new
method and discusses the obtained results. Finally, Section~\ref{sec:conclusion}
provides a summary of the findings and future research directions.
\section{Partitioning Neural Networks}
\label{sec:partition}

This section is divided in three parts. Section~\ref{sec:partition:motivation}
further clarifies the problem which is being solved and how the solution is
related to existing methods in the literature.
Section~\ref{sec:partition:method} describes the method itself, and
Section~\ref{sec:partition:analysis} analyzes the possible benefits achievable
by the method, being some of them confirmed in the experiments performed.

\subsection{Motivation}
\label{sec:partition:motivation}
Large neural networks are able to achieve better performance than smaller ones,
but are considerably more expensive to learn and use. They may require special
methods for training if they do not fit in a single computing unit
\citep{dean2012large} and can be used after trained to provide guidance to
improve smaller networks \citep{Hinton15}. Therefore, even if a large neural
network will not be used for the desired task due to its high computational
cost, training large networks is still important for guiding the improvement of
smaller ones.

Another common solution to improve performance achievable by smaller neural
networks is through ensembles \citep{hansen1990neural}, where predictions of
multiple neural networks are combined to provide more accurate outputs. It is
important to highlight that ensembles only work because the multiple models
provide diverse predictions for the same data \citep{perrone1992networks}.
Therefore, the improvement in performance relies on the distinct behavior of the
predictors.

Since the performance achieved by a neural network may depend on its
initialization, there has been a search for good initialization methods
\citep{glorot2010understanding,sutskever2013importance,he2015delving}.
Nonetheless, neural networks seem to be able to achieve good and diverse local
minima or saddle points
\citep{dauphin2014identifying,choromanska2014loss,pascanu2014saddle}, so they
can easily be used as components of ensembles to improve performance. In the
experiment with the MNIST data set, we will show that the sub-models in fact
achieve diverse performance even when they are small, which indicates that they
learn non-redundant features and provides variety in the features provided as
initial condition for the merged model.

But if we view an ensemble of trained neural networks as a single large
neural network with a constraint of no connection between layers of different
base networks, it raises the question of whether it is possible to use smaller
networks to properly initialize larger ones.
In the initial learning phase, isolated sets of parameters are going to be
adjusted independently, thus avoiding the necessity of using methods devoted to
train the full network with all adjustable parameters.
\subsection{Partitioning Method}
\label{sec:partition:method}

Consider a neural network described by a directed graph from the input to the
output that performs some computation as information flows through the graph.
The case of undirected neural networks will be discussed later, and it will be
shown that they do not modify the algorithm very much.

The proposed partitioning method first divides the neurons in the large neural
network in disjoint sets, except for neurons in the inputs and outputs layers,
which must be present in all sets. In this partition, one filter of a
convolutional layer corresponds to an atomic unit, since all the computed
activations share the same parameters, and any layer that has internal
parameters or whose activation depends on the individual input values instead of
their aggregate, such as normalization layers \citep{krizhevsky2012imagenet},
must be replaced by multiple similar, parallel layers.

For each set of neurons, the only sources and sinks in the vertex-induced
subgraph must be the input and output neurons of the full neural network,
respectively. Therefore, the following holds: 1) each subset defines a complete
flow of information from the input to the output; 2) each vertex-induced
subgraph defines a valid smaller neural network; 3) the original cost function
can be learnt on each subgraph; 4) every neuron of the full neural network is
allocated to exactly one of the smaller networks, except for the input and
output neurons; and 5) every parameter is allocated to at most one smaller
network, except for the parameters of the output neurons.

\begin{figure*}[h]
\centering
\begin{tikzpicture}[scale=0.8, every node/.style={scale=0.8}]
  \node[textneuron] (x) {$X$};

  \node[neuron] (n111) [above right=0.45cm and 1.5cm of x] {};
  \node[neuron] (n112) [above right=0.3cm and 1.5cm of x] {};
  \node[box, inner sep=1mm, fit= (n111) (n112),label=$L_{11}$] (L11) {};

  \node[neuron] (n121) [below right=0.45cm and 1.5cm of x] {};
  \node[neuron] (n122) [below right=0.3cm and 1.5cm of x] {};
  \node[box, inner sep=1mm, fit= (n121) (n122),label=below:$L_{12}$] (L12) {};

  \node[neuron] (n211) [right=2cm of n111] {};
  \node[neuron] (n212) [right=2cm of n112] {};
  \node[box, inner sep=1mm, fit= (n211) (n212),label=$L_{21}$] (L21) {};

  \node[neuron] (n221) [right=2cm of n121] {};
  \node[neuron] (n222) [right=2cm of n122] {};
  \node[box, inner sep=1mm, fit= (n221) (n222),label=below:$L_{22}$] (L22) {};

  \node[textneuron] (y) [below right=0.3cm and 1.5cm of n212] {$Y$};

  \draw[->] (x) edge [left] node [above left=0cm and -0.3cm] {$W_x^{11}$} (L11);
  \draw[->] (x) edge [left] node [below left=0cm and -0.3cm] {$W_x^{12}$} (L12);
  \draw[->] (L11) edge node [above] {$W_{11}^{21}$} (L21);
  \draw[->] (L12) edge node [below] {$W_{12}^{22}$} (L22);
  \draw[->] (L11) edge[dashed] node [below right=-0.1cm and 0.4cm] {$W_{11}^{22}$}
  (L22);
  \draw[->] (L12) edge[dashed] node [below left=-0.1cm and 0.4cm] {$W_{12}^{21}$}
  (L21);
  \draw[->] (L21) edge [right] node [above right=0cm and -0.3cm] {$W_{21}^y$} (y);
  \draw[->] (L22) edge [right] node [below right=0cm and -0.3cm] {$W_{22}^y$} (y);

  \node[textneuron] (x1p) [above right=0.2cm and 1.5cm of y] {$X$};
  \node[neuron] (n111p) [above right=-0.14cm and 1.5cm of x1p] {};
  \node[neuron] (n112p) [below right=-0.14cm and 1.5cm of x1p] {};
  \node[box, inner sep=1mm, fit= (n111p) (n112p),label=$L_{11}$] (L11p) {};
  \node[neuron] (n211p) [right=1.5cm of n111p] {};
  \node[neuron] (n212p) [right=1.5cm of n112p] {};
  \node[box, inner sep=1mm, fit= (n211p) (n212p),label=$L_{21}$] (L21p) {};
  \node[textneuron] (y1p) [below right=-0.13cm and 1.5cm of n211p] {$Y$};
  \draw[->] (x1p) edge node [above] {$W_{x}^{11}$} (L11p);
  \draw[->] (L11p) edge node [above] {$W_{11}^{21}$} (L21p);
  \draw[->] (L21p) edge node [above] {$\hat W_{21}^{y}$} (y1p);

  \node[textneuron] (x2p) [below right=0.4cm and 1.5cm of y] {$X$};
  \node[neuron] (n121p) [above right=-0.14cm and 1.5cm of x2p] {};
  \node[neuron] (n122p) [below right=-0.14cm and 1.5cm of x2p] {};
  \node[box, inner sep=1mm, fit= (n121p) (n122p),label=$L_{12}$] (L12p) {};
  \node[neuron] (n221p) [right=1.5cm of n121p] {};
  \node[neuron] (n222p) [right=1.5cm of n122p] {};
  \node[box, inner sep=1mm, fit= (n221p) (n222p),label=$L_{22}$] (L22p) {};
  \node[textneuron] (y2p) [below right=-0.13cm and 1.5cm of n221p] {$Y$};
  \draw[->] (x2p) edge node [above] {$W_{x}^{12}$} (L12p);
  \draw[->] (L12p) edge node [above] {$W_{12}^{22}$} (L22p);
  \draw[->] (L22p) edge node [above] {$\hat W_{22}^{y}$} (y2p);
\end{tikzpicture}
\caption{Partitioning the full neural network on the left in two sets of neurons
  $S_1 = X \cup L_{11} \cup L_{21} \cup Y$ and $S_2 = X \cup L_{12} \cup L_{22}
  \cup Y$ induces the neural networks on the right. All parameters from the
  smaller models are copied, except for those associated with discarded
  connections, which are zeroed, and those associated with the output, which are
  normalized.}
\label{fig:split}
\end{figure*}
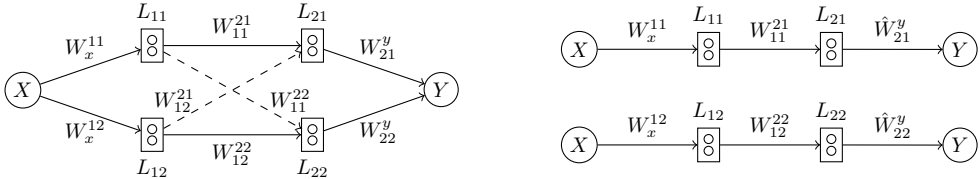

As an example, consider the neural network on the left of
Figure~\ref{fig:split}, which will be separated into two sub-models. The neurons
are grouped in the sets $S_1 = X \cup L_{11} \cup L_{21} \cup Y$ and $S_2 = X
\cup L_{12} \cup L_{22} \cup Y$, which satisfy the conditions imposed before.
When defining the vertex-induced subgraphs, which are shown on the right of
Figure~\ref{fig:split}, the connections between $L_{11}$ and $L_{22}$ and
between $L_{12}$ and $L_{21}$ are not allocated to any subgraph. The extension
for larger number of sub-models is straightforward.

Since the subgraphs are neural networks by themselves and have the same input
and output as the original neural network, the next step in the method is to
train these smaller models independently in the original task. Moreover, since
there is no communication among these networks, they can be trained in parallel.
Once done, almost all the weights and biases can be copied from the sub-models
to the original model directly, except for the output parameters, while setting
to zero all parameters associated with edges that were not trained.

As each sub-model learns to predict the output by themselves, simply copying the
weights would directly sum the contributions of each sub-model, besides each one
providing different values for the biases. While the sum of contributions would
not be a problem for classification problems, since the softmax is
scale-independent, it is a problem for regressions tasks. Therefore, to
normalize the contributions, each weight to the output is divided by the number
K of subtasks created and the output bias is given by the mean of the learnt
biases  of all K sub-models, that is,
\begin{equation}
\label{eq:parameter_adjustment}
  b^y = \frac{\sum_{i=1}^K b_{i}^y}{K}, \quad
  W_{ij}^y = \frac{\hat W_{ij}^y}{K},
  \\
\end{equation}
where $b_i^y, i=1,\ldots,K,$ are the output biases learned by each sub-model.
This change on the parameters is equivalent to computing the mean of the
sub-models on linear outputs or the geometric mean on softmax outputs, which are
the common methods to compute an ensemble prediction. Therefore, the full
network has a behavior similar to the ensemble of the sub-models right after
merging but it is able to learn better parameters as the training continues, as
it is composed of a full network and has more connections than would be present
in the ensemble.

An alternative to changing the weights to the output layer, which would be
necessary in the case of undirected neural networks, since changing the weights
towards the output affect the activation of the hidden units from the input, is
to sum the biases and change the input/output activation function. Instead of
scaling the weights and biases for the output layer, we just have to scale the
input of the functions that describe how the data is generated, as higher
activations will occur. Since the internal parts of the neural network were
trained independently, they can be connected as described before.

Once the parameters of the full model are adjusted based on those from the
sub-models, the fine-tunning can occur by training the full pre-trained model.
\subsection{Analysis of the Method}
\label{sec:partition:analysis}
From the description of the partitioning method, it is clear that each sub-model
of the network will have less parameters to train. After the partition, there
are three possibilities for changing the number of parameters: 1) keep the same
number as the original network, which happens to the output biases; 2) reduce
linearly, which happens to all other biases and input and output weights, since
the number of neurons reduces linearly and the input and output are copied; and
3) reduce quadratically, which happens to the weights between internal neurons.

Since most of the parameters in deep neural networks are concentrated in
connections between internal layers, we can expect an almost quadratic reduction
in the number of parameters in relation to the number of partitions. This
indicates that even a distributed, large neural network can quickly become small
enough to fit in a single machine, which completely eliminates the communication
cost between nodes during the initial training phase. Moreover, it is possible
to further partition a network that is already small enough to fit a single
machine, which reduces the number of operations required for the training and
may further speedup the process.

This not only decreases the communication overhead but can also reduce the
number of machines required during the initial phase of the training. Consider,
for instance, a large neural network that requires four machines connected to be
represented. Due to the space overhead for loading values from other computers,
it might be possible that partitioning the network in two sub-models, each with
about a quarter the original size, allows one of the reduced models to fit in a
single machine. So, besides not requiring communication between the original
machines to train the reduced neural networks, only two of the four original
machines are required for this training step, one for each sub-model of the full
model, supposing that these sub-models are going to be trained in parallel.

Furthermore, we conjecture that the pre-training should not affect the final
performance of the neural network if the partitions have different initial
conditions and are flexible enough. As discussed in
Section~\ref{sec:partition:motivation}, the improved performance achieved by an
ensemble relies on each of its models producing diverse predictions. Moreover,
these diverse predictions may be achieved by distinct representations of the
data in the internal nodes of the neural network.

In this case, after merging the sub-models, new lower-level features for the
data are available to each neuron that was learned in one of the sub-models,
which allows them to exploit these new features to provide better outputs. Since
the new connections have zero value, the network does not have to revert a bad
initial value for the parameters and can directly take advantage of the
promising representations. Moreover, since the representations were also learned
from the data, they should provide useful features and the performance should
not be affected by the smaller parametrization during the pre-training.

\section{Experimental Results}
\label{sec:experiment}
A GeForce GTX 760 was used in these experiments and they were conducted on
neural networks small enough to fit the memory of the GPU, including the
parameters, intermediary values and data batch, but large enough to prevent all
the layer-wise processing to be done completely in parallel. This avoids
overheads due to communication between nodes that may be specific to the method
used to perform model parallelism, while allowing the proposed method to show
its improvements.

We present two experiments to evaluate the method proposed in this paper. The
first uses a small network to classify digits on MNIST \citep{lecun1998gradient}
and focuses on analysing the effects of the number of partitions created on the
training time and performance. The second uses a larger network to classify the
images on CIFAR10 \citep{krizhevsky2009learning} and considers different
number of epochs for the pre-training.

\subsection{MNIST}
\label{sec:experiment:mnist}
For this task, we used a neural network similar to LeNet
\citep{lecun1998gradient}, which was composed of three layers, all with ReLU
activation. The first two layers are convolutions with 20 and 50 filters,
respectively, of size 5x5, both followed by max-pooling of size 2x2, while the
last layer is fully connected and composed of 500 hidden units with dropout
probability of 0.5. The learning was performed by gradient descent with learning
rate of 0.1 and momentum of 0.9, which were selected using the validation set to
provide the best performance for the full model without pre-training, and
minibatches of 500 samples. When the parameters of the sub-models are copied to
the full neural network, the accumulated momenta of each training task are also
copied to the respective sub-set of the full network and adjusted like the
associated parameters. Hence, the accumulated momenta for the weights and biases
for the output layer are computed using Eq.~\eqref{eq:parameter_adjustment} and
the other accumulated momenta are copied directly.

The neural network was partitioned in 2, 5 and 10 sub-models with the same
number of neurons in each layer, besides the baseline of 1 sub-model, which is
the standard neural network model. Each sub-model was trained for 100 epochs on
the full data set, followed by another 100 epochs on the full model with the
merged parameters. The performance of each sub-model was averaged to obtain the
results during pre-training, as they are considered independent models and do
not belong to an ensemble. We also compared the performance of merging the
sub-models before any pre-training, which is equivalent to setting the initial
values for the inter-model weights to zero. However, since this approach
achieved slightly worse results than the baseline, the results are not shown.

Tests with different numbers of sub-models used the same initialization for the
weights, avoiding fluctuation of the results due to random number generation. A
total of 50 runs with different initialization were performed and the averages
of the runs are reported.

\begin{table}[t]
\caption{Comparison between networks with different numbers of sub-models
  against the baseline given by K = 1. The speedup is measured over one
  iteration of one sub-model.}
\label{tab:results:mnist}
\vspace{-0.1in}
\begin{center}
\begin{tabular}{ccc}
\# Sub-models (K) & \# Parameters & Speedup \\
\hline
1 & 431080 (100\%) & 1$\times$ \\
2 & 109295 (25.35\%) & 2.10$\times$ \\
5 & 18224 (4.23\%) & 3.17$\times$ \\
10 & 4867 (1.13\%) & 4.17$\times$ \\
\end{tabular}
\end{center}
\vspace{-0.2in}
\end{table}

Table~\ref{tab:results:mnist} presents the reduction in the number of parameters
and computing time of each pre-training epoch with varying number of sub-models.
As expected by the analysis in Section~\ref{sec:partition:analysis}, the number
of parameters is reduced almost quadratically with the number of sub-models,
significantly decreasing the memory requirements for their storage.
The speedup, on the other hand, presents a diminishing return effect. Although
training on different machines would be faster for all values of K, training on
a single machine would only be faster for K = 2, which provides an argument for
partitioning neural networks even when they fit on memory, as already discussed
in Section~\ref{sec:partition:analysis}. This reduction occurs because the
neural network is relatively small already, so reducing it by a large amount
leaves unused cores of the GPU during training.

\begin{figure}[h]
  \centering
  \hfill
  \begin{subfigure}{0.35\linewidth}
  \psfrag{model_1}[t][c]{\footnotesize Sub-model 1}
  \psfrag{model_2}[b][c]{\footnotesize Sub-model 2}
  \psfrag{val cost}[c][c]{}
  \psfrag{test cost}[c][c]{}
  \includegraphics[width=\linewidth]{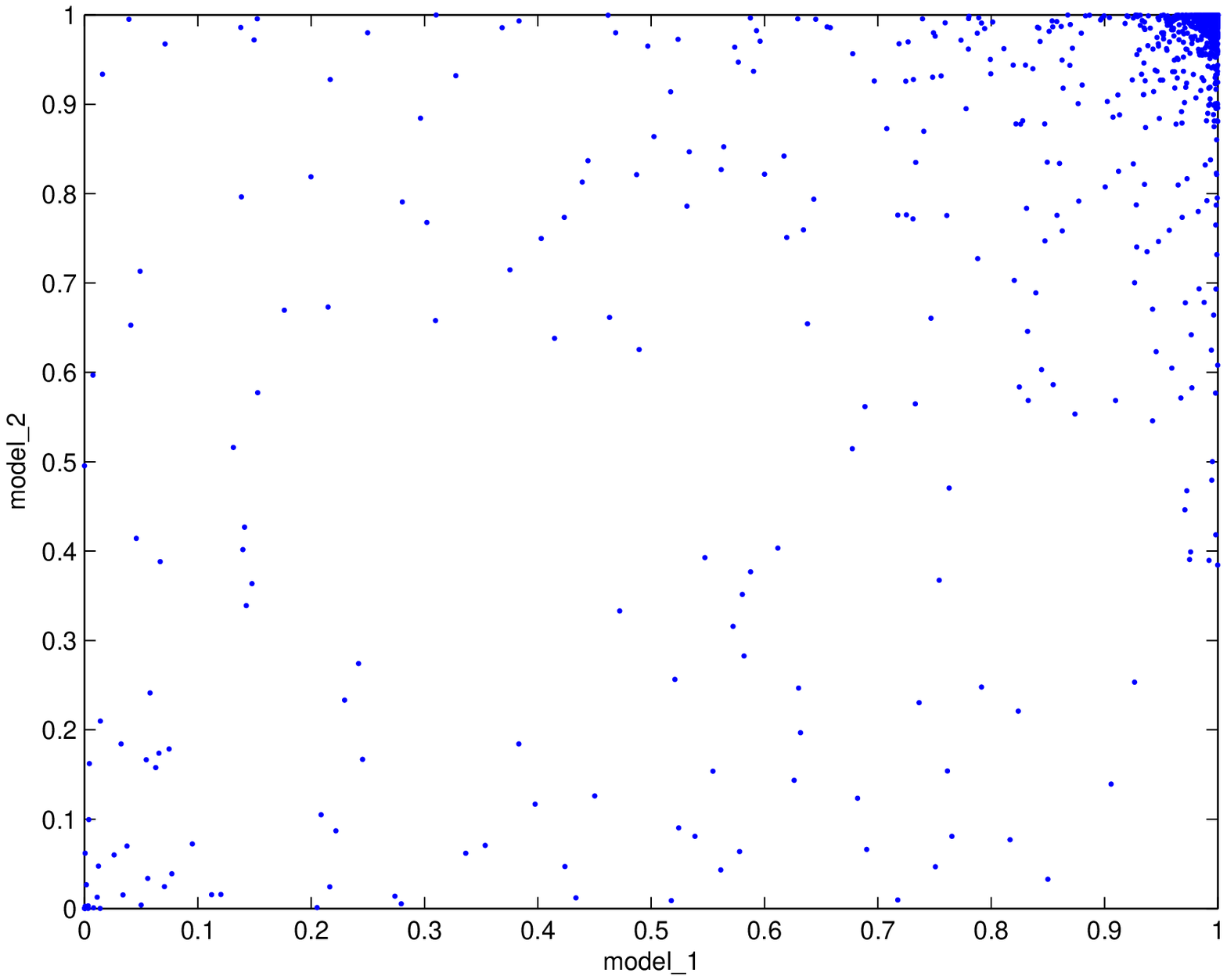}
  \caption{K = 2}
  \end{subfigure}
  \hfill
  \begin{subfigure}{0.35\linewidth}
  \psfrag{t}[t][c]{\footnotesize Time [s]}
  \psfrag{error}[b][c]{\footnotesize Classification errors}
  \psfrag{val errors}[c][c]{}
  \psfrag{test errors}[c][c]{}
  \includegraphics[width=\linewidth]{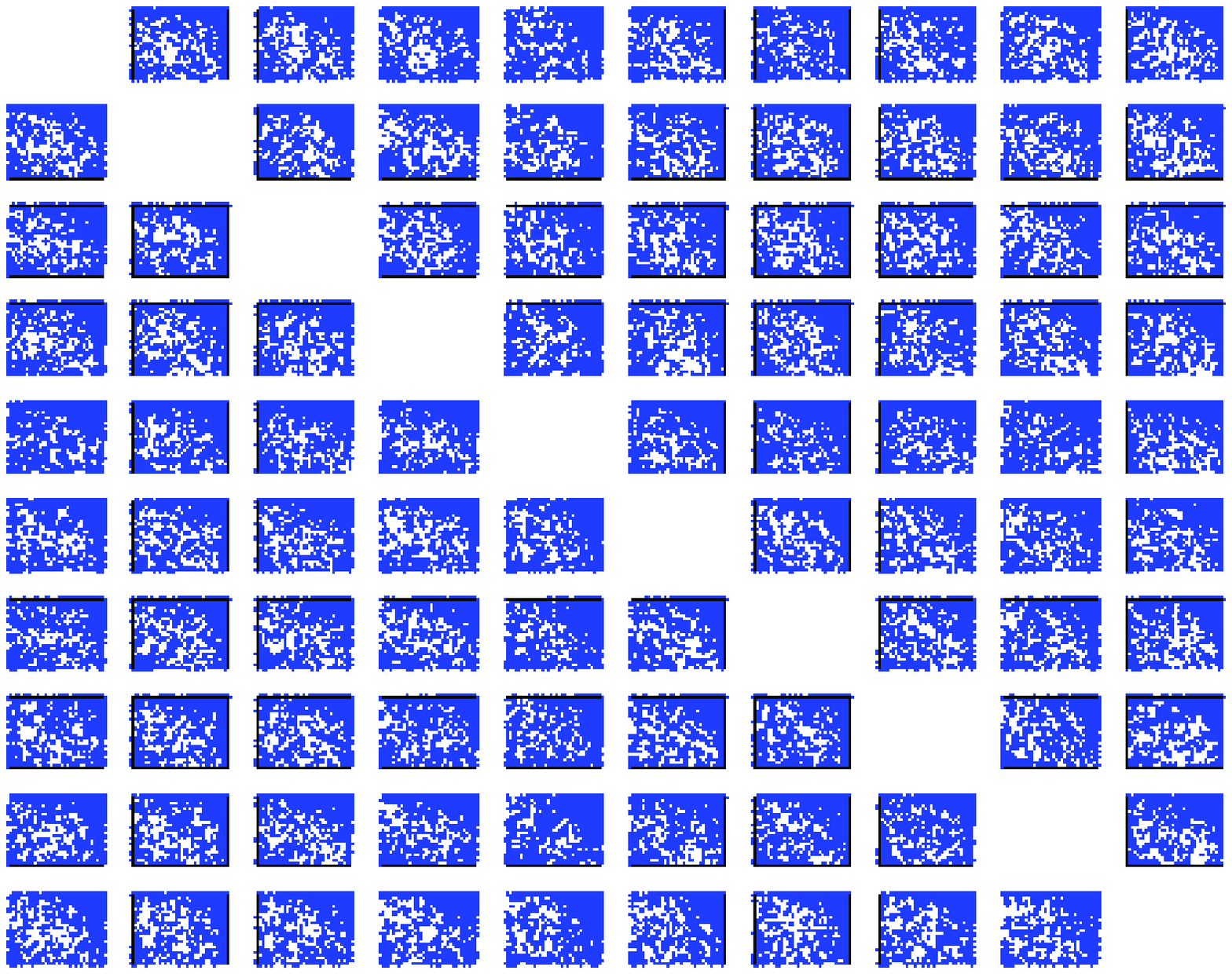}
  \caption{K = 10}
  \end{subfigure}
  \hfill
  \hfill
  \caption{Probability assigned by each sub-model to the correct class over the
  test set for different numbers of sub-models just before merging. For K = 10,
  the probabilities for each pair of sub-models is shown. Though not shown here,
  results for K = 5 are intermediary, thus indicating that diversity increases
  somehow proportional to the number of submodels.}
  \label{fig:predictions}
\end{figure}

\begin{figure}[h]
  \centering
  \hfill
  \begin{subfigure}{0.4\linewidth}
  \psfrag{t}[t][c]{\footnotesize Time [s]}
  \psfrag{error}[b][c]{\footnotesize Mean loss}
  \psfrag{val cost}[c][c]{}
  \psfrag{test cost}[c][c]{}
  \includegraphics[width=\linewidth]{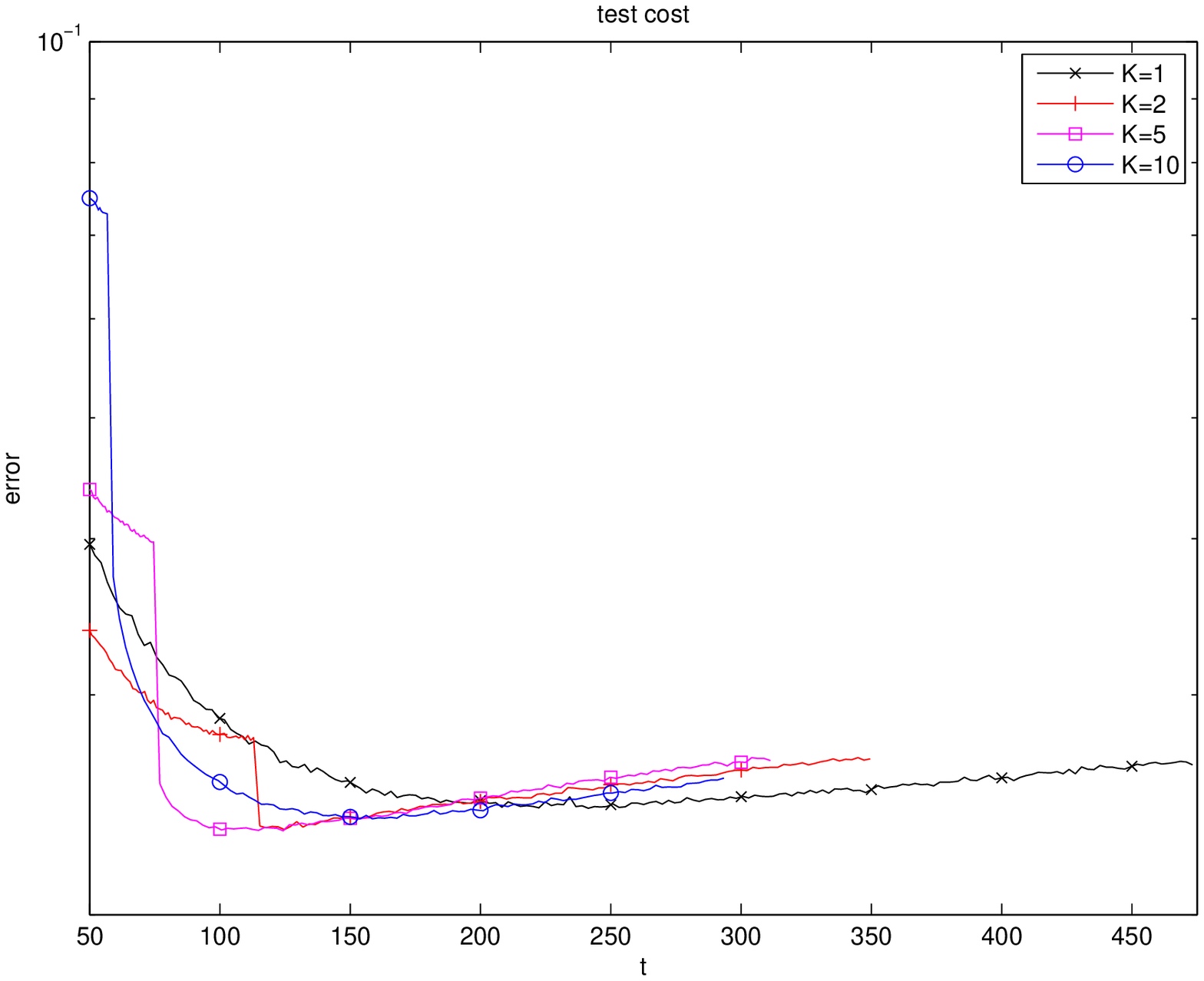}
  \end{subfigure}
  \hfill
  \begin{subfigure}{0.4\linewidth}
  \psfrag{t}[t][c]{\footnotesize Time [s]}
  \psfrag{error}[b][c]{\footnotesize Classification errors}
  \psfrag{val errors}[c][c]{}
  \psfrag{test errors}[c][c]{}
  \includegraphics[width=\linewidth]{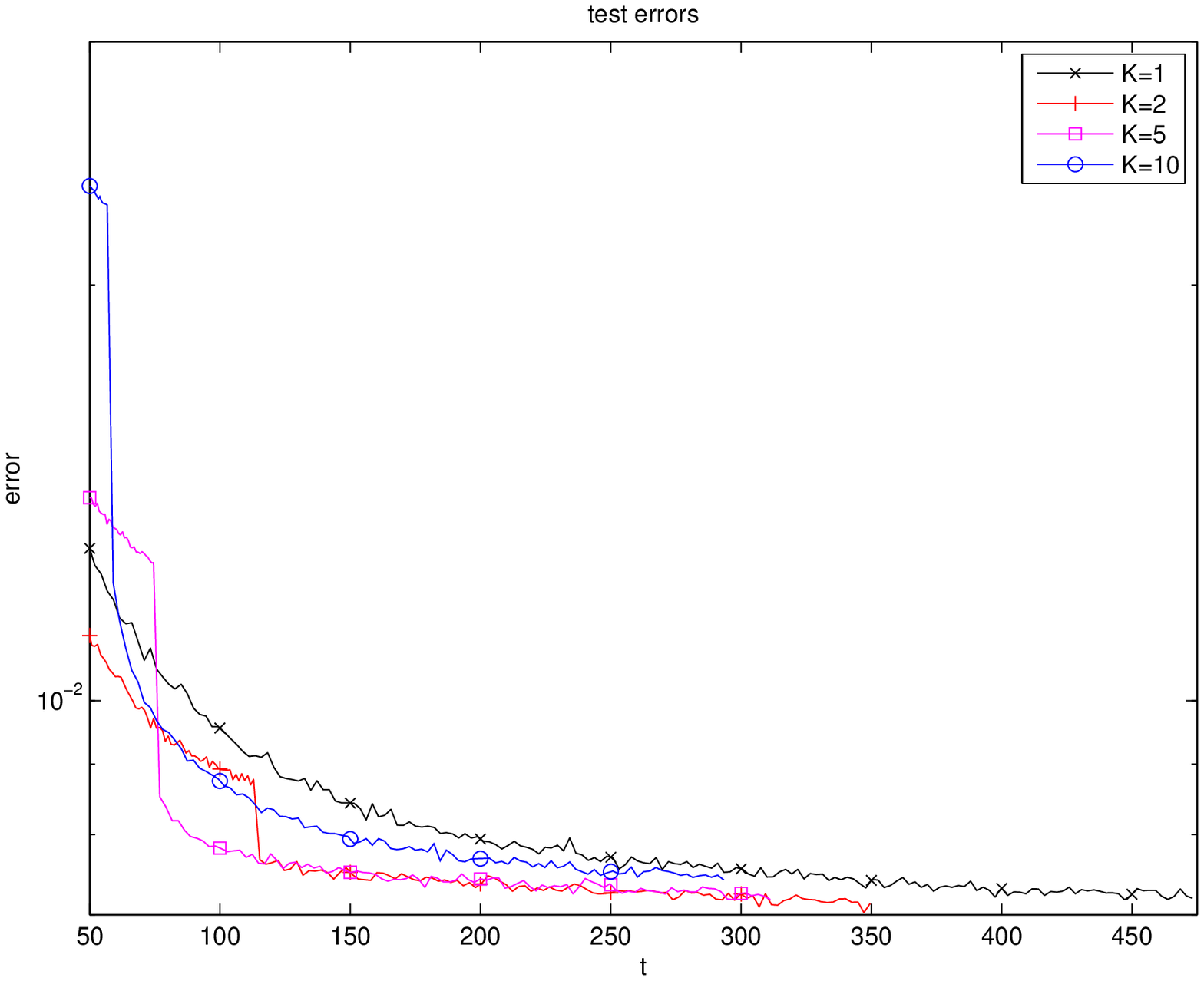}
  \end{subfigure}
  \hfill
  \hfill
  \caption{Mean loss (left) and classification error (right) over the test
  set on a partitioned network with K sub-models, including pre-training and
  training after merging.}
  \label{fig:test}
\end{figure}

As anticipated in Section~\ref{sec:partition:motivation}, it is important that
the sub-models learn different representations and predictions for the data, as
otherwise there would be redundant information provided by the initialization.
To measure the diversity, we evaluated the probability given by each sub-model
to the correct class on each sample in the test set just before merging the
sub-models and the resulting images are shown in Figure~\ref{fig:predictions}.
It is clear that in both cases the sub-models indeed have different predictions
for the data, with more diverse values when smaller sub-models are considered,
and the case for 5 sub-models, not shown, presenting an intermediary diversity.
This result helps to lessen the concern about the size of sub-models used and
provides support that different initializations for each sub-model may be enough
to provide the required diversity. However, another important factor to take
advantage of the sub-models is present in Figure~\ref{fig:test}, which shows
that the sub-models have not converged before merging, since this could have
reduced the diversity present in the sub-models.

Figure~\ref{fig:test} shows the mean loss and classification error for the test
set. It is clear that the pre-training method showed faster convergence to the
same performance than the full training, providing evidence for the conjecture in
Section~\ref{sec:partition:analysis} that the proposed method should not affect
the performance of the network. The merging instant is clearly marked by the
step-like change in the curves. One noticeable result present in this image is
that using 2 or 5 sub-models provided similar performance in similar time
scales, although 5 sub-models have faster iterations. This occurs because,
although the 5 smaller models are able to learn diverse representations, these
representations do not have the same quality level as the ones learned with 2
sub-models due to the flexibility of the sub-models. When merging the 5
sub-models, the learning algorithm has to adjust the pre-trained features more,
requiring more time on the full model. This also explains the curve of
classification error for 10 sub-models, as these models have even less
flexibility and a large part of the learning has to be performed after merging.

It is important to highlight that the speedup achieved was relatively small due
to the size of the network. On larger networks, the benefit of partitioning the
neural network during the first part of the training should be higher, as will
be shown in the next section.
\subsection{CIFAR10}
\label{sec:experiment:cifar}
The neural network and parameters used for these experiments are the same used
by \citet{krizhevsky2012imagenet} in the CUDA-ConvNet
library\footnote{https://github.com/akrizhevsky/cuda-convnet2}, which is
composed of two convolutional layers with ReLU activation, each one followed by
normalization and pooling layers. After that, two locally connected layers
with ReLU activation and a fully connected layer complete the network. The
details of the network and parameters can be found in the source code of the
library.

The neural network has more than 2.1 million parameters and we fix the number of
partitions to 2 and vary the merging iteration. Since the datapath in the
original network presents normalization layers, it is not possible to apply the
pre-training algorithm directly, since these layers would require communication
between the sub-models. To solve this issue, as discussed in
Section~\ref{sec:partition:method}, we duplicate each normalization layer and
connect them to different subsets of the pooling layer, so that a valid
partition of the vertices of the graph can be performed.

We compare the performances and training times of the full model, which works as
the baseline, the full model with duplicated normalization layers, and
pre-trained models merging at different epochs, all trained with the same
parameter setting to optimize the performance of the baseline. This allows us to
evaluate the difference in behavior by just using the proposed pre-training
method in an existing training schedule for the full model. Like in the MNIST
test, the accumulated momenta of each sub-model are copied with their
corresponding weights when merging the sub-models.

\begin{table}[t]
\caption{Comparison between different merging epochs and two versions of the
  full model, with mean value over 20 runs and standard deviation in
  parentheses. The training time and speedup shown correspond to the full
  training and consider that the partitions were trained in series (left) or in
  parallel (right).
  There are 10000 images in the test set and the classification error is the
  number of misclassified images.}
\label{tab:results:cifar}
\vspace{-0.1in}
\begin{center}
\begin{tabular}{cccccc}
\multicolumn{2}{c}{\multirow{2}{*}{Model}} &
Net2WiderNet &
Partitioning &
\multirow{2}{*}{Training time (min)} &
\multirow{2}{*}{Speedup} \\
& & classif. error & classif. error & & \\
\hline
\multicolumn{2}{c}{Original} & \multicolumn{2}{c}{1182.95 (18.69)} & 142.39 & 0.88$\times$\\
\multicolumn{2}{c}{Duplicated} & \multicolumn{2}{c}{1184.60 (19.06)} & 124.69 & 1.00$\times$\\
\multirow{11}{*}{\begin{tabular}{@{}c@{}}Merging\\iteration\end{tabular}}
&   0 & 1175.75 (13.78) & 1177.20 (19.34) & 124.69 & 1.00$\times$\\
&  50 & 1189.80 (23.71) & 1183.25 (20.89) & 120.18 / 116.89 & 1.04$\times$ / 1.07$\times$\\
& 100 & 1187.80 (21.87) & 1180.15 (18.87) & 115.68 / 109.10 & 1.08$\times$ / 1.14$\times$\\
& 150 & 1198.80 (16.52) & 1177.55 (18.36) & 111.18 / 101.31 & 1.12$\times$ / 1.23$\times$\\
& 200 & 1198.85 (24.57) & 1173.75 (12.64) & 106.67 / 93.51 & 1.17$\times$ / 1.33$\times$\\
& 250 & 1200.85 (20.01) & 1178.40 (21.15) & 102.17 / 85.72 & 1.22$\times$ / 1.45$\times$\\
& 300 & 1219.60 (18.34) & 1177.05 (24.13) & 97.67 / 77.93 & 1.28$\times$ / 1.60$\times$\\
& 350 & 1244.05 (18.46) & 1177.55 (19.46) & 93.16 / 70.13 & 1.34$\times$ / 1.78$\times$\\
& 400 & 1278.15 (16.38) & 1184.10 (20.91) & 85.89 / 59.57 & 1.45$\times$ / 2.09$\times$\\
& 450 & 1325.95 (17.23) & 1191.75 (24.87) & 78.62 / 49.01 & 1.59$\times$ / 2.54$\times$\\
& 500 & 1388.50 (12.36) & 1257.70 (18.15) & 71.34 / 38.44 & 1.75$\times$ / 3.24$\times$\\
\end{tabular}
\end{center}
\vspace{-0.2in}
\end{table}

Table~\ref{tab:results:cifar} shows the results for each one of the models
considered. Note that duplicating the normalization layers and dividing inputs
among them, named ``Duplicated'' in the table, decreases the training time
because there are less inputs being considered at each normalization unit,
reducing the time required to compute the normalized data.

For merging epoch up to 400, there is no noticeable trend in the performance of
the classifier, with fluctuations occurring due to the stochastic optimization.
On the other hand, the new pre-training method reduced the training time
significantly even when serial training of the partitions is considered, with
higher improvements assuming parallel training.
Since the training schedule performs 520 iterations, the higher error for the
last two merges can be explained by the optimization algorithm not having enough
time to adjust the new weights created and initially set to zero by the merge.

Each training iteration over the merged model takes about 3.3 seconds while
iterations over one partition takes 1 second, characterizing a speedup of 3.3
times. When compared against the speedup of 2.1 times obtained on the MNIST
network, which was about 5 times smaller, this provides evidence for the
conjecture in Section~\ref{sec:experiment:mnist} that larger networks benefit
more from the pre-training. Therefore, even larger neural networks should have
higher improvements on their training time, since more computations are saved
and less computing power is idle during the training of the sub-models.

Table~\ref{tab:results:cifar} also show the results obtained by applying
Net2WiderNet \citep{net2net} to one of the sub-models to double the size of each
layer, so that the expanded network has the same topology as the network after
the partition, but with different parameters. Instead of the random expansion
presented by \citet{net2net}, we duplicated each layer completely as the new size
is multiple of the original one, and we added noise to the duplicated parameters
to break symmetry, as described by \citet{net2net}\footnote{We experimented with
many ways to set the noise level and to copy the accumulated momenta to the new
network. The best results were obtained when adding to each parameter set a
uniform noise with amplitude of 2\% of the maximum absolute value in the set and
by resetting the momenta to zero after expanding the network.}. Since only one
model is being trained, only the second (higher) speedup applies. From the
results obtained when limiting the number of iterations of the base and expanded
models, it is clear that Net2WiderNet presents a smaller speedup than our
partitioning method for the same level of error. Namely, Net2WiderNet is able to
perform only 100 iterations on the base model, which corresponds to a speedup of
1.14 times, before the mean error gets higher than 1190, while our method is
able to perform 400 iterations on the two sub-models before achieving the same
error level, which corresponds to a speedup of 1.45 times if we consider that
the models are trained serially.

We must highlight that, despite these results, Net2Net may be faster when
training a new model from an existing, pre-trained one, specially if the
expansion is not as large as the one pursued here, and allows changes to the
model that would require a training from scratch using our method, like
increasing the number of neurons in a single layer, since it is not possible to
create a complete sub-model from input to output to append to the existing
network. Therefore, a combination of our partitioning method, to quickly get to
a large base model, and Net2Net methods, to experiment with changes in this base
model, may be the best choice and should be investigated in the future.
\section{Conclusion}
\label{sec:conclusion}
In this paper, we introduced a method for pre-training a neural network by
partitioning it into smaller neural networks and training them on the original
learning task. The size of the sub-models reduces almost quadratically with the
number of sub-models created, which allows larger neural networks to save more
computational resources during pre-training.

By design, the method decreases the training time by creating training subtasks
that can be parallelized and may reduce the communication overhead present in
model parallelism. It may also decrease the number of computing units required
during the pre-training due to the quadratic reduction on the number of
parameters being learned.

Two experiments, on MNIST and on CIFAR10, with neural networks that fit in a
single GPU, confirmed that the training time of the full model can be decreased
without affecting the performance. The experiments also show that the proposed
method may be able to improve training speed even if the training of the
sub-models is performed serially and that larger models may experience higher
speedups, with a speedup on the pre-training iterations of 2.1 for a 3-layer
model with 430k parameters and a speedup of 3.3 for a 4-layer model with 2.1M
parameters when creating 2 sub-models.

Since the proposed method relies on the different sub-models finding diverse
representations for the data that can be exploited once they are merged, it is
plausible to worry about the size of the sub-models created, as smaller models
have less flexibility and might learn similar functions. However, the MNIST
experiment shows that indeed the sub-models learn diverse representations for
the data, with the pairwise diversity being higher when we increase the number
of sub-models.

Future research should experiment with deep neural networks using both data and
model parallelism to evaluate the gains obtained by reducing the models. The
time spent training the neural network can be decomposed mainly in three parts,
all of which may be affected by the proposed method: 1) the time to compute the
gradients and adjust the parameters, which we showed that can achieve large
improvement; 2) the time to communicate updates between units with different
parts of the data, which is proportional to the number of parameters and should
get an almost quadratic speedup; and 3) the time to communicate activations
between computers when using model parallelism, which can be decreased or even
avoided by using smaller models. These reductions in training time and the
possibility of reducing the number of computers required during pre-training, as
discussed in Section~\ref{sec:partition:analysis}, make the proposed method
appealing to handle large models.

Another direction is the evaluation of a hierarchical training method, in which
the neural network is decomposed recursively and the pre-training is performed
at each level, such that the proposed method characterizes one level of
recursion. This could provide additional benefits to very large models by
allowing smaller sub-models to be trained without merging all of them at the
same time.

\subsubsection*{Acknowledgments}
The authors would like to thank CNPq for the financial support.

\small{\bibliography{paper}}
\bibliographystyle{templates/iclr/iclr2016_stylefiles_conference/iclr2016_conference}

\end{document}